%% file: main.tex
\documentclass[a4paper]{article}

\usepackage[a4paper]{geometry}

\usepackage[utf8]{inputenc}
\usepackage[T1]{fontenc}
\usepackage{hyperref}
\usepackage{url}

\usepackage{booktabs}
\usepackage{nicefrac}
 \usepackage{microtype}
\usepackage{pifont}

\usepackage{graphicx}
\usepackage{subcaption}

\usepackage{tabularx,ragged2e}
\usepackage{multirow}

\usepackage{authblk}

\usepackage{mathtools,empheq,amsfonts,amsthm,amsmath,amssymb,bm,mathrsfs}

\usepackage{appendix}

\usepackage{xcolor}
\usepackage{anyfontsize}
\usepackage{cleveref}
\usepackage{soul}

\usepackage{algorithm}
\usepackage{algpseudocode}

\usepackage[round]{natbib}
\usepackage{todonotes}

\newcommand{\R}{\mathbb R}

\newcommand{\softmax}{\mathrm{softmax}}
\newcommand{\softplus}{\mathrm{softplus}}
\newcommand{\fft}{\mathrm{FFT}}
\newcommand{\ifft}{\mathrm{IFFT}}
\newcommand{\sign}{\mathrm{sign}}
\newcommand{\roll}{\mathrm{roll}}
\newcommand{\pad}{\mathrm{pad}}
\newcommand{\logapprox}{\widetilde \log}

\algnewcommand\algorithmicparfor{\textbf{parallel for}}
\algdef{SE}[PAR]{ParFor}{EndParFor}[1]%
  {\algorithmicparfor\ #1 \algorithmicdo}%
  {\algorithmicend\ \algorithmicparfor}

\title{Parallelizable Neural Turing Machines}

\author[1]{Gabriel Faria}
\author[2]{Arnaldo Candido Junior}
\affil[1]{University of São Paulo\footnote{Instituto de Ciências Matemáticas e de Computação (ICMC), University of São Paulo.}}
\affil[2]{São Paulo State University\vspace{4pt}}
\affil[ ]{{ \texttt{gabriel.om.faria@usp.br}, \texttt{arnaldo.candido@unesp.br} }}
\date{}

\begin{document}

\maketitle

\begin{abstract}
    We introduce a parallelizable simplification of Neural Turing Machine~(NTM), referred to as P-NTM, which redesigns the core operations of the original architecture to enable efficient scan-based parallel execution.
    We evaluate the proposed architecture on a synthetic benchmark of algorithmic problems involving state tracking, memorization, and basic arithmetic, solved via autoregressive decoding.
    We compare it against a revisited stable implementation of the standard NTM, as well as conventional recurrent and attention-based architectures.
    Results show that, despite its simplifications, the proposed model attains length generalization performance comparable to the original, learning to solve all problems, including unseen sequence lengths, with perfect accuracy.
    It also improves training efficiency, with parallel execution of P-NTM being up to an order of magnitude faster than the standard NTM.
    Ultimately, this work contributes toward the development of efficient neural architectures capable of expressing a broad class of algorithms.
\end{abstract}

\input{sections/01_introduction}
\input{sections/02_ntm}
\input{sections/03_par_ntm}
\input{sections/04_experiments}
\input{sections/05_discussion}
\input{sections/06_limitations}

\input{sections/07_related_work}
\input{sections/08_conclusion}

\bibliography{main.bib}

\clearpage

\appendix

\input{appendix/a}
\input{appendix/b}
\input{appendix/c}

\end{document}

%% file: sections/01_introduction.tex
\section{Introduction}
\label{sec:introduction}

Memory-augmented neural networks offer an alternative to traditional sequence models by integrating explicit, structured memory components.
These architectures are typically inspired by classical models of computation and seek to make neural networks more expressive by giving them computational capabilities similar to their theoretical counterparts.

Such networks commonly employ memory structures such as stacks~\citep{das1992stack,joulin2015stack}, queues~\citep{grefenstette2015transduce}, tapes~\citep{graves2014ntm}, among other mechanisms.
However, despite their increased expressiveness, these approaches are generally far less efficient than modern language modeling architectures, such as Transformers~\citep{vaswani2017attention} and state-space models like Mamba~\citep{gu2024mamba,dao2024mamba2}.
In particular, their reliance on inherently sequential memory operations makes them difficult to parallelize, significantly limiting their scalability and practical applicability to large-scale training settings.

In response to this, we revisit the Neural Turing Machine (NTM) architecture~\citep{graves2014ntm}.
Motivated by recent advances in developing parallelizable recurrent networks~\citep[e.g.,][]{feng2024rnns}, we investigate how NTM can be adapted for parallel execution.
In this process, we provide two contributions:
\begin{itemize}
    \item We reexamine the original NTM architecture, reassessing its implementation and initialization hyperparameters, and propose targeted modifications that improve stability and performance (Section \ref{sec:ntm}).
    \item We introduce the {Parallelizable Neural Turing Machine}~(P-NTM) architecture\footnote{An implementation is available at \url{https://github.com/gbrlfaria/pntm}.}, which simplifies several aspects of the original architecture, enabling the use of \emph{parallel scan} algorithms for efficient parallelization and effective processing of long input sequences; and we explain how, despite these simplifications, the new architecture can represent computations equivalent to those of the original architecture (Section \ref{sec:parntm}).
\end{itemize}

We then empirically validate the proposed architecture on a benchmark of synthetic algorithmic tasks, evaluating the ability of each architecture to learn and express different kinds of computation, as well as the performance gains enabled by parallelization (Section \ref{sec:experiments}).
We show that the proposed model matches the original architecture in generalization performance, with both consistently generalizing to substantially longer, unseen sequence lengths, while providing significant speedup gains enabled by parallelization (Section \ref{sec:results}).
We further discuss these findings in Section \ref{sec:discussion}.
Finally, we outline limitations of the proposed architecture and directions for future work (Section \ref{sec:limitations}) and situate our contributions in relation to prior work (Section \ref{sec:relwork}).

%% file: sections/02_ntm.tex
\section{Neural Turing Machines}
\label{sec:ntm}

\begin{figure}
	\centering
	\includegraphics[width=0.6\linewidth]{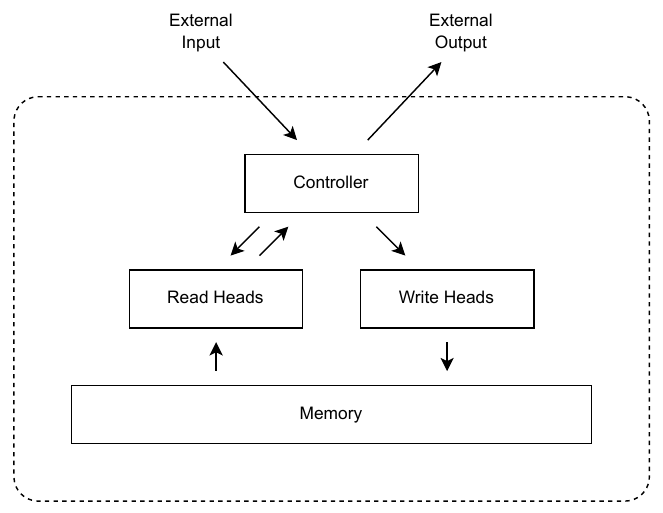}
	\caption{Illustration of the NTM architecture. Adapted from \citet{graves2014ntm}.}
	\label{fig:ntm}
\end{figure}

NTM~\citep{graves2014ntm} is a neural network architecture inspired by the classical Turing machine, consisting of a learnable neural controller that interacts with an external memory matrix through fully differentiable read and write operations, which are carried out by a set of read and write heads.
Using this mechanism, it can successively interact with memory to perform structured computation over input sequences and represent a variety of algorithmic patterns.
An illustration of the architecture is provided in \Cref{fig:ntm}.

At each time step, the controller receives the current input along with past information, which includes content retrieved from memory and, in the case of recurrent controllers, a hidden state.
Based on the combined input, the controller generates control signals that direct the read and write heads.
Some of these signals handle memory addressing, determining where to read from or write to, while others specify the content to be written and erased.
After executing the read and write operations, the NTM model generates an output based on its updated internal state.

Below, we present our definition of the NTM architecture and elaborate on its components, with particular emphasis on the modifications introduced in our implementation.
This serves both to clearly establish the baseline used in this paper and as a preliminary to our main contribution.

\subsection{Definition}
\label{subsec:ntm}

An NTM can be defined as a sequence-to-sequence mapping that transforms an input sequence of vectors $\bm{x}_1, \ldots, \bm{x}_T$, where each $\bm{x}_t \in \mathbb{R}^d$, into an output sequence $\bm{y}_t, \ldots, \bm{y}_T$, where each $\bm{y}_t \in \mathbb{R}^d$.
This mapping is characterized by the following recurrent update equation, applied at each time step $t$:
\begin{equation}
\label{eq:ntm}
(\bm y_t, \bm h_t, \bm M_t, \bm A^r_t, \bm A^w_t, \bm r_t) = \mathrm{NTM}(\bm x_t, \bm h_{t-1}, \bm M_{t-1}, \bm A^r_{t-1}, \bm A^w_{t-1}, \bm r_{t-1}),
\end{equation}
where $\bm h_t \in \mathbb{R}^c$ is the internal state of the controller\footnote{We assume a general class of recurrent models that can be described using a single hidden state vector.} and $\bm M_t \in \mathbb{R}^{m \times n}$ is the memory matrix consisting of $m$ cells\footnote{In general, the number of memory cells need not be fixed and can be dynamically adjusted to match the size of the problem at hand.} of dimension $n$, both at time $t$.
Meanwhile,
$\bm A_{t}^{r} = (\bm a^{r}_{t,1}, \ldots, \bm a^{r}_{t,H})$ and $\bm A_{t}^{w} = (\bm a^{w}_{t,1}, \ldots, \bm a^{w}_{t,H})$
respectively denote the addressing weights for each of the $H$ read and write heads\footnote{As a simplifying assumption, we enforce an equal number of read and write heads.} (hence the $r$ and $w$ superscripts) at time~$t$.
Each vector $\bm{a}_{t,h}^{r}, \bm{a}_{t,h}^{w} \in [0,1]^m$ represents a weighting over the $m$ memory locations, with entries summing to one.
These vectors specify how strongly the $h$th read or write head focuses on each memory location.
Finally, $\bm r_t \in \mathbb{R}^{Hn}$ is the read vector obtained by aggregating the outputs of the $H$ read heads at time $t$.

The recurrent update equation~\eqref{eq:ntm} that defines the NTM is governed by its control, addressing, memory writing, memory reading, and output mechanisms.
Each of these components is formally presented in the following sections.
For clarity, we describe the behavior of each read and write head in isolation, noting extensions to the multi-head setting where appropriate.

\subsubsection{Control}

The control mechanism begins by updating the controller based on its previous state, the current input, and the read vector from the previous step.
Formally, given a controller function $f_c$, the update is expressed as:
\begin{align}
    \label{eq:controller}
    \bm h_t = f_c(\bm x_t, \bm r_{t-1}, \bm h_{t-1}),
\end{align}
where the initial read vector is set to $\bm r_0 = \bm 0$ by default.
The controller function $f_c$ may be instantiated as any suitable neural network layer\footnote{For example, an RNN, an LSTM, or even a feedforward network (if the previous hidden state is disregarded).}.
The resulting controller state $\bm h_t$ is then used to compute the parameters that guide the operations of the NTM at the current step.

For each read or write head, the following set of vectors and scalars is produced to control the addressing mechanism, specifying which memory locations the head will focus on:
\begin{align}
    \bm k_t &= \tanh ( \bm W_{k} \bm h_t + \bm b_k), \\
    \beta_t &= \softplus ( \bm w_{\beta} \bm h_t + b_\beta ), \\
    g_t &= \sigma ( \bm w_{g} \bm h_t + b_g ), \\
    \bm s_t &= \softmax ( \bm W_{s} \bm h_t + \bm b_s ), \\
    \gamma_t &= 1 + \softplus ( \bm w_{\gamma} \bm h_t + b_\gamma ),
\end{align}
where $\bm k_t \in (-1, 1)^n$ is the key vector, $\beta_t > 0$ is the key strength, $0 < g_t < 1$ is the interpolation gate, $\bm s_t \in (0,1)^3$ is the shift vector, and $\gamma_t > 1$ is the sharpening factor.
Meanwhile, $\bm W_{k} \in \R^{n \times c}$, $\bm W_{s} \in \R^{3 \times c}, \bm w_{\beta}, \bm w_{g}, \bm w_{\gamma} \in \R^{1 \times c}$ and $\bm b_{k} \in \R^{n}$, $b_{s} \in \R^{3}, b_{\beta}, b_{g}, b_{\gamma} \in \R^1$ are learnable parameters specific to each head.
The precise role of each addressing control will be explained in the following sections.

Additionally, for each write head, the following two control vectors are computed to determine how the memory matrix is updated:
\begin{align}
    \bm u_t &= \tanh(\bm W_{u} \bm h_t + \bm b_u), \\
    \bm d_t &= \sigma(\bm W_{d} \bm h_t + \bm b_d),
\end{align}
where $\bm u_t \in (-1,1)^n$ is the add (or update) vector, $\bm d_t \in (0,1)^n$ is the  erase (or delete) vector, and
$\bm W_u, \bm W_d \in \R^{n \times c}$ and $\bm b_u, \bm b_d \in \R^n$ are learnable parameters specific to each write head.
Again, the specific roles of these vectors will be detailed in the subsequent sections.

\subsubsection{Addressing}

Addressing operates through two primary mechanisms.
The first is content-based addressing, which functions similarly to an attention mechanism by focusing on memory cells whose contents closely match a given key vector.
The second is location-based addressing, which allows the head to shift to adjacent memory locations, enabling sequential memory access patterns similar to those of a classical Turing machine.
These two mechanisms are combined to produce a final vector of addressing weights.
The details of this process are outlined below.

\paragraph{Initialization.}
Before processing begins, addressing weights must be initialized to set the initial focus of the heads.
While \citet{graves2014ntm} did not prescribe a specific initialization scheme, subsequent work proposed learning them as a parameter vector~\citep{collier2018ntm}.
Instead, we initialize all addressing weights to focus solely on the first memory cell: $\bm{a}_0 = [
\begin{matrix}
	1 & 0 & \cdots & 0
\end{matrix}].$

\paragraph{Content-based addressing.} Given a similarity function $K: \mathbb{R}^n \times \mathbb{R}^n \to \mathbb{R}$ between vectors, typically instantiated as cosine similarity, the content-based weights $\bm{a}^c_t$ are computed by measuring the similarity between the current key vector $\bm{k}_t$ emitted by the controller and each memory cell~$\bm{M}_t[i]$, for $i = 1, \dots, m$:
\begin{equation}
    \bm{a}_t^c[i] = \frac{\exp(\beta_t \mathop{K} ( \bm{k}_t, \bm{M}_t[i] ) )}{\sum_{j=1}^m \exp(\beta_t \mathop{K}(\bm{k}_t, \bm{M}_t[j]))},
\end{equation}
where the key strength $\beta_t$ adjusts the sharpness of the weight distribution, controlling how strongly similarity affects the weights.

\paragraph{Interpolation.}
Before proceeding, the newly computed content-based weights are combined with the previous weights using the current interpolation gate $g_t$, yielding the interpolated weights:
\begin{equation}
   \bm a^g_t = (1-g_t) \bm a_{t-1} + g_t \bm a^{c}_t.
\end{equation}
This interpolation allows the model to smoothly transition between relying on previous addressing weights and using the new content-based addressing.

\paragraph{Location-based addressing.} The interpolated weights are then shifted left, right, or kept unchanged according to the shift vector $\bm{s}_t$ output by the controller at the current time step.
This operation yields the shifted weights $\bm{a}^s_t$, computed via the circular convolution:
\begin{equation}
    \label{eq:ntm-conv}
    \bm a_t^s[i] = \sum_{j \in \{-1,0,1\}} \bm a^g_t[{i-j}]\, \bm s_t[j],
\end{equation}
where $\bm{a}^g_t[i - j]$ denotes\footnote{The index $i - j$ is treated cyclically over the range $1$ to $m$. For example, if $i - j = 0$, the index wraps to $m$; if $i - j = m + 1$, it wraps to 1.} the interpolated addressing weight at offset $-j$ from position $i$ and $\bm{s}_t[j]$ is the shift strength associated with offset $j$ (i.e., shift left by 1, no shift, or shift right by 1).
This allows the weights to smoothly shift across time steps, wrapping around the memory matrix at its boundaries.

\paragraph{Sharpening.}
Because shifts are not perfectly discrete, their repeated application leads to a blurring effect, where the addressing weights increasingly dissipate.
To counteract this, a final sharpening step is applied using the sharpening factor $\gamma_t$, producing the final addressing weights~$\bm a_t$.
This process can be expressed in two equivalent forms:
\begin{equation}
    \bm a_{t}[i]
    \stackrel{(a)}{=} \frac{(\bm a_t^s[i])^{\gamma_t}}{\sum_{j=1}^m (\bm a_t^s[j])^{\gamma_t}}
    \stackrel{(b)}{=} \exp \Big (\gamma_t \log \bm a_t^s[i] - \log \Big ( \sum_{j=1}^m \exp (\gamma_t \log \bm a_t^s[j] ) \Big) \Big).
\end{equation}
Crucially, the original formulation (a) of the sharpening step is numerically unstable: moderately large exponents can zero out entries and lead to divide-by-zero errors, leading to \texttt{NaN} errors.
While prior work has attempted to address this by clipping the outputs of the controller to an arbitrarily chosen range~\citep{collier2018ntm}, we instead resolve the issue by adopting formulation (b), implemented with the log-sum-exp trick for numerical stability.

\subsubsection{Memory Writing}

Once the addressing weights for the current step have been computed, the model writes to memory in two steps: first, it selectively erases existing content; then, it adds new content.
The erase vector $\bm{d}_t$ specifies which elements of the memory cells should be cleared, specifically at the locations attended to by the write head.
Next, the add vector $\bm{u}_t$ determines the new content to be added to those same locations.
Formally, the update of each memory cell is defined as:
\begin{equation}
\label{eq:ntm-memory}
    \bm{M}_t[i] = (\bm{1} - \bm a_{t}[i]\, \bm{d}_t )\odot \bm{M}_{t-1}[i] + \bm a_{t}[i]\, \bm{u}_{t},
\end{equation}
where $\odot$ denotes elementwise vector multiplication.

When multiple write heads are used, all erasures from the different heads are applied first, followed by all corresponding additions.
Let $\bm{d}_{t,1}, \dots, \bm{d}_{t,H}$ and $\bm{u}_{t,1}, \dots, \bm{u}_{t,H}$ denote the erase and add vectors from each of the $H$ write heads.
Then, the multi-head memory update takes the same form as~\eqref{eq:ntm-memory}, with a multiplicative erase term $\prod_{h=1}^{H} (\bm{1} - \bm{a}^w_{t,h}[i]\, \bm{d}_{t,h})$ and an additive term~$\sum_{h=1}^{H} \bm{a}^w_{t,h}[i]\, \bm{u}_{t,h}$, for each memory location $i = 1, \dots, m$.

For initialization, we follow the approach of \citet{collier2018ntm} where each element of the memory matrix is initialized to a small positive constant by default.

\subsubsection{Memory Reading}
After writing has taken place, the model reads from memory by computing a weighted sum of the updated memory cells using the addressing weights of the read head.
This results in the read vector:
\begin{equation}
    \bm{r}_t = \sum_{i=1}^m \bm a_{t}[i] \bm{M}_t[i],
\end{equation}
which captures the relevant content from the memory matrix.
When multiple read heads are employed, each head independently generates its own read vector by applying its respective addressing weights to memory.
These individual read vectors are then concatenated to form the final read vector $\bm r_t$.

\subsubsection{Output}

Finally, the model produces an output vector $\bm{y}_t$ based on the current controller state and read vector, effectively combining information from the current input as well as both current and past memory.
This is computed as:
\begin{equation}
    \bm{y}_t = \bm{W}_{o,h} \bm{h}_{t} + \bm W_{o,r} \bm{r}_{t} + \bm b_o,
\end{equation}
where $\bm W_{o,h} \in \R^{d \times c}, \bm W_{o,r} \in \R^{d \times Hn}, \bm b_o \in \R^d$ are learnable parameters.

%% file: sections/03_par_ntm.tex
\section{Parallelizable Neural Turing Machines}
\label{sec:parntm}

P-NTM is a simplified and parallelizable variant of the standard NTM.
It retains the central concept of a recurrent model with multiple read and write heads operating over a memory matrix, but introduces key simplifications primarily aimed at enabling efficient parallel computation.

Most notably, it forgoes content-based addressing and eliminates temporal dependencies in the control mechanism.
That is, head controls at each time step depend solely on the current input and are unaffected by past memory or control signals.
As a result, this design enables internal states to be computed in parallel across time steps using scan operations.

In the following sections, we present the P-NTM architecture in detail.
We begin by introducing its recurrent formulation~(\Cref{subsec:parntm}) and then explain how its parallel mode of execution~(\Cref{subsec:parallel-implementation}).
Next, we present a stabilization mechanism aimed at improving inference over long sequences~(\Cref{subsec:stabilization}).
Finally, we discuss how P-NTM can represent computation and contrast it with the standard NTM (\Cref{subsec:computation}).

\subsection{Recurrent Formulation}
\label{subsec:parntm}

A P-NTM can be formulated analogously to the standard NTM described in \Cref{subsec:ntm}, but with a simplified update equation:
\begin{equation}
(\bm y_t, \bm M_t, \bm A^{r}_t, \bm A^{w}_t) = \mathrm{P\text{-}NTM}(\bm x_t, \bm M_{t-1}, \bm A^{r}_{t-1}, \bm A^{w}_{t-1}),
\end{equation}
where dependencies on the recurrent controller state and the previous read vector are removed, so that the model maintains only the memory matrix and two sets of addressing weights for the read and write heads.

Specifically, at each step, the model computes head controls based on the current input.
These controls determine how the read and write heads shift their focus over the memory matrix as well as what content should be written to memory.
Writes are non-interfering, with each write head targeting a distinct portion of each memory cell.
A learnable mixing mechanism allows each read head to access content from any write head.
The final output is computed by projecting the aggregated readings from the read heads.

We now describe the components of this recurrent update in detail.
As before, we consider a single pair of read and write heads in isolation, noting extensions to the multi-head scenario where applicable.
The full multi-head implementation is provided in the appendix.

\subsubsection{Control}

For each pair of read and write heads, the control mechanism generates three components based on the inputs: a read shift vector, a write shift vector, and a memory update vector.
These are respectively defined as:
\begin{align}
	\bm s^{r}_t & = \softmax (\bm W_{r} \bm x_t), \\
	\bm s^{w}_t & = \softmax (\bm W_{w} \bm x_t), \\
	\bm u_t     & = \bm W_{u} \bm x_t,
\end{align}
where $\bm W_{r}, \bm W_{w} \in \R^{3 \times d}$ and $\bm
W_{u} \in \R^{n \times d}$ are learnable parameters specific to each pair of heads.
The shift vectors $\bm s^{r}_t, \bm s^w_t \in (0,1)^3$ represent the strength of different head shift operations (i.e., left, stay, right), while the update vector $\bm u_t \in \R^n$ specifies the new content to be written to memory in the current step.

\subsubsection{Addressing}

The addressing mechanism follows the location-based strategy of the standard NTM.
By design, the initial weights are fully concentrated on the first memory cell: $\bm{a}_0= [
\begin{matrix}
	1 & 0 & \cdots & 0
\end{matrix}].$
At each subsequent time step, the addressing weights of each head are updated via circular convolution using the previous weights and the current shift vector.
For each memory location $i = 1, \dots, m$:
\begin{equation}
	\bm a_t[i] = \sum_{j \in \{-1,0,1\}} \bm a_{t-1}[i-j]\, \bm s_t[j],
\end{equation}
where $\bm a_{t-1}[i - j]$ is indexed cyclically and $\bm s_t[j]$ denotes the shift strength for offset $j$, as in \eqref{eq:ntm-conv}.

A subtle but important detail is that both reading and writing operations in P-NTM rely on the addressing weights $\bm a_{t-1}$ from the previous time step, rather than the current weights $\bm a_t$.
As a result, the addressing weights are typically updated as the final operation in each recurrent computation step.
This design choice reflects the behavior of classical Turing machines, where reading and writing occur at the location left by the previous transition.

\subsubsection{Memory Writing}

The memory matrix is first initialized to zeros: $\bm M_0 = \bm 0$.
At each time step, the contents of each memory cell are updated using the previous write addressing weights and the current update vector:
\begin{equation}
    \label{eq:par-mem-update}
    \bm M_t[i] = (1 - \bm a_{t - 1}[i]) \bm M_{t-1}[i] + \bm a_{t - 1}[i]\, g(\bm u_t),
\end{equation}
where $g$ is a nonlinearity introduced by \cite{feng2024rnns}, defined elementwise as:
\begin{equation}
g(x) =
    \begin{cases}
        x + 0.5, & \text{if } x \geq 0 \\
        \sigma(x), & \text{otherwise}.
    \end{cases}
\end{equation}
This nonlinearity ensures that the memory contents remain differentiable, non-negative, and unbounded above.
These characteristics, in turn, enable an effective log-space representation of memory, a property that will become important later.

After the update, each memory cell is mixed through a linear projection:
\begin{equation}
\label{eq:mixing}
    \bm{\tilde M}_{t}[i] = \bm W_{m} \bm{M}_t[i],
\end{equation}
where $\bm W_{m} \in \mathbb{R}^{n \times n}$ is a learnable matrix.

In the multi-head setting, each memory cell is partitioned among multiple write heads.
For instance, with a 12-dimensional cell and 3 write heads, dimensions 1--4 go to head 1, 5--8 to head 2, and 9--12 to head 3.
This arrangement ensures that each head writes to a distinct portion of the memory cell, avoiding interference between them.
Yet, because of the projection in~\eqref{eq:mixing}, information written by a single head is distributed across the entire memory cell, allowing interaction between the contributions of different heads.

\subsubsection{Memory Reading}

After writing, the model retrieves information from memory by computing a weighted sum over the mixed memory cells using the addressing weights of the read head from the previous step, producing the read vector:
\begin{equation}
    \bm r_t = \sum_{i=1}^{m} \bm a_{t - 1}[i] \, \bm{\tilde M}_t[i].
\end{equation}
With multiple read heads, each head performs a separate read using its own addressing weights, and their read vectors are concatenated to form the final vector $\bm r_t$.

\subsubsection{Output}

The final output at each time step is computed by linearly transforming the read vector:
\begin{equation}
\bm y_t = \bm W_{o} \bm r_t,
\end{equation}
where $\bm W_{o} \in \mathbb{R}^{d \times Hn}$ is a learnable output projection matrix.
In the multi-head setting, the concatenated read vectors from all heads are jointly transformed by this matrix, enabling the model to integrate information from multiple heads into the final output.

\subsection{Parallel Implementation}
\label{subsec:parallel-implementation}

\begin{figure}[t]
	\centering
	\includegraphics[width=\linewidth]{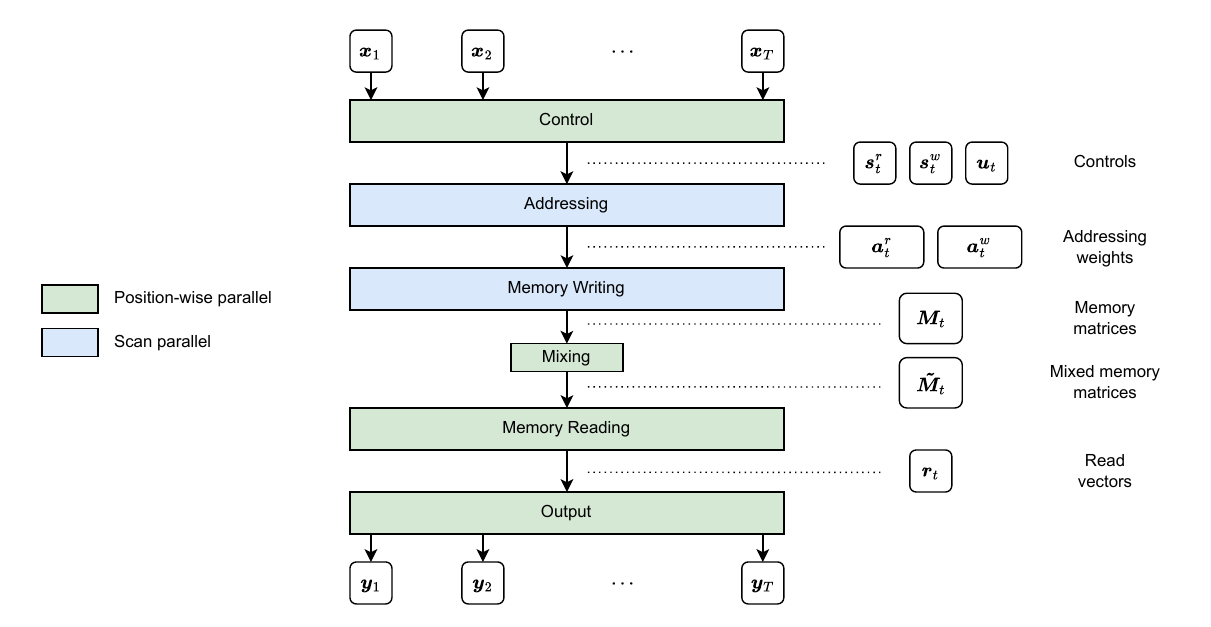}
	\caption{Illustration of the parallel computation of P-NTM.}
	\label{fig:parallel-parntm}
\end{figure}

In the parallel execution of P-NTM, computation proceeds in a few well-defined stages.
First, all input vectors are processed simultaneously to produce the shift and update control vectors.
The shift controls are then converted into read and write addressing vectors using a parallel scan operation.
Next, given the write addresses and update vectors, another scan generates the memory states for all sequence positions.
Finally, the read vectors are computed in parallel from the memory states and read addresses, followed by the construction of the output sequence.

Notably, parallelism in P-NTM arises in two main forms.
Position-wise parallelism occurs when operations at each sequence position, such as computing the control vectors, are independent and can be executed simultaneously.
Scan parallelism handles the recurrences in addressing and memory updates by reformulating them as prefix computations, allowing cumulative operations to be efficiently calculated across the sequence.
The overall structure of this parallel computation strategy is illustrated in \Cref{fig:parallel-parntm}.

Below, we introduce the scan-based primitive used in P-NTM and explain how it enables parallel computation of addressing weights and memory states across time steps.

\subsubsection{Parallel Scan}

The parallel scan algorithm~\citep{blelloch1990prefix} enables the efficient computation of all prefix values of a sequence, obtained by repeatedly applying an associative binary operator $\oplus$ such as addition or multiplication.
Given a sequence of inputs $u_1, u_2, \ldots, u_T$, the scan operation produces the prefix sequence:
\begin{equation}
    u_{1}, (u_{1} \oplus u_2), \ldots, (u_1 \oplus u_2 \oplus \cdots \oplus u_T).
\end{equation}
Because of the associativity of the binary operator, the computation can be structured as a tree, allowing it to be distributed across multiple processors.
Relative to sequence length, this parallelization achieves linear speedup with respect to the number of processors, introducing only logarithmic overhead.

The scan operation serves as a basic primitive for a broad class of computations, including cumulative sums, cumulative products, and, generally, first-order recurrences of the form:
\begin{equation}
\label{eq:recurrence}
    v_t= \alpha_t v_{t-1} + \beta_{t},
\end{equation}
where $\alpha_{t}$ and $\beta_{t}$ are real-valued scalars.
Given the initial value $v_{0}$ and the sequences $\alpha_{1}, \dots, \alpha_{T}$ and $\beta_{1}, \dots, \beta_{T}$, the scan algorithm can compute the entire sequence $v_{1}, \dots, v_{T}$ in parallel.
As we will see, these computations form the basis of both the addressing and memory update operations in P-NTM, which in turn enables them to be executed in parallel.

\subsubsection{Shift-based Addressing}

As previously mentioned, updating addressing weights using a shift vector from one time step to the next corresponds to a circular convolution operation.
This perspective enables us to leverage the mathematical properties of convolution to efficiently compute sequences of shifts, allowing for parallel construction of addressing vectors across time steps.

According to the convolution theorem~\citep{oppenheim1999signal}, the circular convolution between two sequences (or vectors) $\bm x$ and $\bm y$ can be computed through an elementwise product in the frequency domain using the fast Fourier transform (FFT) and its inverse (IFFT):
\begin{equation}
\label{eq:conv}
    \bm x \ast \bm y = \ifft(\fft(\bm x) \odot \fft(\bm y)).
\end{equation}

We use this fact to compute addressing in parallel.
Let $\bm{k}_t \in \mathbb{R}^m$ be the kernel corresponding to the shift $\bm{s}_t$, obtained by padding the shift vector to the memory size $m$ and circularly shifting its elements to the left.
Then, the addressing vector at time $t$ can be computed as:
\begin{align}
    \bm a_t    & = \ifft (\bm \phi_t), \\
    \bm \phi_t & = \fft (\bm a_0) \odot \fft (\bm k_1) \odot \cdots \odot \fft (\bm k_{t-1}).
\end{align}
This formulation enables the computation of addressing weights at each time step by applying an inverse Fourier transform to the cumulative elementwise product of the transformed kernels.
Importantly, as discussed in the previous section, this product admits a parallel implementation.

However, this method can suffer from numerical underflow due to the repeated multiplication of small-magnitude values.
Since both addressing weights and shift vectors have entries less than or equal to one, their corresponding frequency-domain representations may quickly decay when multiplied over many time steps, leading to vanishingly small values.

To address this, we revisit the convolution computation.
Rather than performing the operation directly, we compute the convolution in log-space by replacing elementwise multiplications with sums of approximate complex logarithms.
The resulting approximation is given by:
\begin{equation}
	\bm x \ast \bm y \approx \ifft (\exp (\logapprox (\fft (\bm x)) + \logapprox (\fft (\bm y)))),
\end{equation}
where the approximate logarithm is applied elementwise as:
\begin{equation}
	\logapprox (x) =
	\begin{cases}
		\log (x + \varepsilon \cdot \sign (x)), & \text{if } x \ne 0 \\
		\log (\varepsilon),                     & \text{otherwise}.
	\end{cases}
\end{equation}
Here, $\varepsilon$ is a small positive constant and the complex $\sign$ function denotes the unit vector in the direction of $x$.
This approximation avoids the undefined behavior of the complex logarithm at zero and mitigates the instability caused by large gradients near zero.

Ultimately, this approach enables a stable approximation of each addressing vector through a cumulative sum that can be parallelized across the sequence length, such that:
\begin{align}
	\bm a_t            & \approx \ifft (\exp(\bm{\tilde \phi}_t)),                                                                     \\
	\bm{\tilde \phi}_t & = \logapprox (\fft (\bm a_0)) + \logapprox (\fft (\bm k_{1})) + \cdots + \logapprox (\fft (\bm k_{t-1})).
\end{align}
The complete implementation is detailed in \Cref{alg:conv}, which includes an additional clamping step to account for approximation errors in the addressing weights, keeping them within the valid range of 0 to 1.

\begin{algorithm}[t]
	\caption{$ \bm a_0, \dots, \bm a_{T-1} \gets \texttt{ConvShift}(\bm s_{1}
	, \dots, \bm s_{T-1})$}
	\label{alg:conv}
	\begin{algorithmic}
		[1] \Statex \textbf{Input:} $\bm s_1, \dots, \bm s_{T-1}$, sequence of
		shift vectors, where each $\bm s_t \in\mathbb{R}^{3}$. \Statex \textbf{Output:}
		$\bm a_{0}, \dots, \bm a_{T-1}$, sequence of addressing vectors, where
		each $\bm a_t \in \mathbb{R}^{m}$.

		\ParFor {$t \gets 0\textbf{ to }T-1$}
            \State
		$\bm k_t \gets \roll_{\text{-}1}(\pad_{m}(\bm s_t)) \textbf
		{ if }t > 0 \textbf{ else }[1, 0, \dots, 0]$\footnotemark
		\State $\bm \psi_t \gets \logapprox (\fft (\bm k_t))$
		\EndParFor

		\State $\bm{\tilde \phi}_{0}, \dots, \bm{\tilde \phi}_{T-1} \gets \mathrm{cumsum}
		( \bm \psi_{0}, \dots, \bm \psi_{T-1})$

		\ParFor {$t \gets 0\textbf{ to }T-1$} 
            \State
		$\bm a_t\gets \ifft (\exp(\bm{\tilde \phi}_t))$ \State
		$\bm a_t\gets \mathrm{clamp}_{[0,1]}(\bm a_t)$
            \EndParFor

		\State \Return $\bm a_{0}, \dots, \bm a_{T-1}$
	\end{algorithmic}
\end{algorithm}

\subsubsection{Memory Writes}

From the memory update rule in \eqref{eq:par-mem-update}, each scalar entry of the memory matrix evolves according to a first-order recurrence, as defined in \eqref{eq:recurrence}, with $v_0 = 0$, $\alpha_{t} = 1 - \bm a_{t-1}[i]$, and $\beta_t = a_{t-1}[i]\, g(\bm u_t[j])$, for each memory cell $i$ and component $j$.
Because all $\alpha_t$ and $\beta_t$ are known in advance (since all addressing and update vectors are known at this point), the parallel scan algorithm can be applied to compute memory updates efficiently over the entire sequence.

To ensure numerical stability and maintain precision over long sequences, we follow the method used by \citet{feng2024rnns}, which builds on the log-space parallel scan algorithm of \citet{heinsen2023scan}.
This approach computes the recurrence using cumulative sums in log-space, allowing for efficient and stable evaluation.
In our case, the log-space representation is approximate, as exact zeros in the addressing weights cannot be expressed.
Nonetheless, it remains an effective implementation.

The complete parallel memory writing procedure is described in \Cref{alg:mem}.
Write addressing weights and their complements are broadcast across memory cells to enable elementwise updates.
After parallel updates are applied, each memory cell is linearly projected as described in \eqref{eq:mixing}.

\begin{algorithm}[t]
	\caption{$\bm M_1, \dots, \bm M_t \gets \texttt{MemoryWrite}(\bm a_0
        \label{alg:mem}
	, \dots, \bm a_{T-1}, \bm{u}_1, \dots, \bm{ u}_t)$}
	\begin{algorithmic}
		[1] \Statex \textbf{Input:} $\bm a_0, \dots, \bm a_{T-1}$, sequence of
		addressing vectors, where each $\bm a_t\in \R^{m}$; \Statex \phantom{\textbf{Input:}}
		$\bm{u}_1, \dots, \bm{u}_t$, sequence of update vectors, where each
		$\bm{u}_t\in \mathbb{R}^{n}$. \Statex \textbf{Output:} $\bm M_1, \dots
		, \bm M_t$, sequence of memory matrices, where each $\bm M_t\in \mathbb{R}
		^{m \times n}$.

		\ParFor {$t \gets 1\textbf{ to }T$} \Comment{Compute scan coefficients.}
                \State $\bm \alpha_t\gets \log(\mathrm{clamp} 
		_{[\varepsilon, 1-\varepsilon]}(\bm a_{t-1}))$
                \State $\bm B_t\gets \log
		(1 - \exp(\bm \alpha_t)) + \log g(\bm{u}_t)^{\intercal}$ \EndParFor 

		\State
		$\bm \alpha^{\star}_{1}, \dots, \bm \alpha^{\star}_t \gets \mathrm{cumsum}
		(\bm \alpha_1, \dots, \bm \alpha_t)$ \Comment{Apply scan.}
		\State
		$\bm H_1, \dots, \bm H_t \gets \mathrm{cumlogsumexp}(\bm B_{1}- \bm
		\alpha^{\star}_{1}, \dots, \bm B_t- \bm \alpha^{\star}_t)$
		\State
		$\bm M_{1}, \dots, \bm M_t \gets \exp(\bm \alpha_1^{\star} + \bm
		H_{1}), \dots, \exp(\bm \alpha_t^{\star} + \bm H_t)$

		\State \Return $\bm M_{1}, \dots, \bm M_t$
	\end{algorithmic}
\end{algorithm}

\footnotetext{Here, $\mathrm{pad}_m(\cdot)$ extends the 3-element shift vector to length $m$ by appending zeros, and $\mathrm{roll}_{-1}(\cdot)$ circularly shifts the vector by $-1$ (i.e., one position to the left). This ensures that the FFT-based circular convolution applies the shift kernel according to the intended left-stay-right weighting.}

\subsection{Stable Inference}
\label{subsec:stabilization}

As discussed earlier, because shifts are generated by a softmax and thus not perfectly discrete, their repeated application leads to a blurring effect, causing the addressing weights to progressively dissipate.
While the standard NTM addresses this issue using a learnable sharpening mechanism, we instead apply a separate stabilization step during inference.

Specifically, any shift strength below a threshold hyperparameter $0 \leq \tau < 1$ is set to zero.
The remaining shift strengths, which exceed $\tau$, are then recomputed by applying the softmax only over their corresponding offsets.
The rationale is that we may interpret extremely small shift weight values as a result of the mathematical inability of the softmax function to produce exact zeros, combined with a difficulty of the optimizer in converging toward such distributions.
Therefore, sufficiently small values of $\tau$ may allow discrete shift patterns to be effectively represented by suppressing shift directions with negligible weights.

Finally, we note that inference is performed recurrently in linear space, following the equations presented in \Cref{subsec:parntm}, rather than the log-space parallel implementation described in \Cref{subsec:parallel-implementation}.
This is so that discrete behavior can be properly represented.

\subsection{Computational Mechanism}
\label{subsec:computation}

Standard NTMs perform computation via a nonlinear recurrent controller that maintains a hidden state across time steps, serving as the finite-state control mechanism analogous to that of a classical Turing machine.
At each step, the controller combines the current input, information retrieved from memory, and its previous hidden state to generate control signals governing read and write operations.
Continuous nonlinear recurrent networks of the form used in NTMs, as specified in \Cref{eq:controller}, are known to be capable of representing finite automata~\citep{siegelmann1996finite,weiss2018practical}.
This capability enables the controller to realize the finite-state component of a Turing machine, while the memory matrix provides an analog of the tape.

P-NTM, by contrast, departs from this design by eliminating the explicit controller state.
All control decisions are generated solely from the current input, with no direct dependence on memory contents or recurrent hidden states.
This architectural choice prevents the model from tracking the finite-state component of computation via an internal state register, as in standard NTMs, and therefore requires computation to be realized through an alternative mechanism.

Instead, P-NTM achieves structured computation through autoregression.
The model consumes its own outputs as subsequent inputs while continuing to read from and write to the memory matrix.
In this setting, the current input token (typically the most recent output) acts as an implicit state variable that conditions control at each step.
State information is therefore encoded in the output sequence itself, rather than in a persistent hidden vector, effectively distributing state management across time and eliminating the need for a nonlinear recurrent controller.
The following sections present empirical experiments and results that illustrate the computational capabilities of this autoregressive formulation.

%% file: sections/04_experiments.tex
\section{Experiments}
\label{sec:experiments}

We evaluate the proposed P-NTM architecture in two ways.
First, we empirically assess its ability to learn and represent various algorithmic behaviors in an autoregressive setting (i.e., via \emph{chain of thought}), despite the simplifications introduced relative to the original design.
Second, we verify the computational efficiency gains enabled by its parallelism compared to the original architecture.
The general experimental setup for each evaluation is described in the following sections, with further details being provided in Appendix~\ref{ap:experimental-details}.

\subsection{Algorithm Learning and Representation}

The experimental design for assessing learning and representation of algorithms follows that of \citet{deletang2023chomsky}.
It consists of training models with different architectures on various algorithmic tasks (e.g., evaluating an arithmetic expression) and determining their ability to generalize to larger, previously unseen instances of those tasks.
A model is considered to have successfully learned the underlying algorithm if it can consistently solve instances larger than those seen during training.
The rationale for these tasks is to simulate, in a controlled and simplified context, the reasoning and computational challenges typically found in real-world problems, which are often expressed in natural language and addressed by large language models.

\begin{table}
    \centering
    \caption{Algorithmic tasks with input and output examples. Final answers are underlined. Detailed descriptions of each task are provided in Appendix~\ref{ap:experimental-details}.}
    \begin{tabular}{lll}
        \hline
        \textbf{Task} & \textbf{Example Input} & \textbf{Example Output} \\
        \hline
        Parity Check (PC) & \( aaabba \) & \( 01000\underline{1} \) \\
        Cycle Navigation (CN) & \(siidis\) &  \( 01212\underline{2} \) \\

        Reverse String (RS) & \( aabba \) & \( \underline{abbaa} \) \\
        Duplicate String (DS) & \(aabba\) & \( \underline{aabbaaabba} \) \\

        Modular Arithmetic (MA) & \( 1 + 2-4 \) & \( {+}10{+}21{-}43\underline{4} \) \\
        Binary Addition (BA) & \( 01101 + 101 \) & \( \underline{11011} \)  \\
        \hline
    \end{tabular}
    \label{tab:problems}
\end{table}

\paragraph{Tasks.}
Each algorithmic task consists of taking an input sequence of tokens, which represents a particular problem instance, followed by a special separator token, and producing an output sequence of tokens representing its solution.
The output is generated step by step, including all intermediate tokens required to reach the final answer.
The tasks incorporated into our experiments are a subset of those from \citet{deletang2023chomsky}, further modified to include explicit intermediate steps for every problem.
They are intended to probe computational capabilities such as state-tracking, memorization, and basic arithmetic.
\Cref{tab:problems} summarizes the tasks and provides representative input and output examples.

\paragraph{Architectures.}
Several neural network architectures are evaluated in the experiment, including not only the neural Turing machines central to this work but also standard sequence models.
This broader comparison provides clearer context for interpreting the results.
Specifically, both standard recurrent and parallelizable linear recurrent architectures are included as baselines, using LSTM~\citep{hochreiter1997lstm} and minimal gated recurrent unit (minGRU)~\citep{feng2024rnns} networks, respectively.
Autoregressive Transformers~\citep{vaswani2017attention} are also assessed under different position encoding schemes, including NoPE (no positional encoding), ALiBi~\citep{press2022alibi}, RoPE~\citep{su2024rope}, and FIRE~\citep{li2024fire}.
The evaluation further includes a standard NTM with an LSTM controller and a P-NTM preceded by a minGRU layer.
In this configuration, the minGRU serves as a parallelizable recurrent aggregator of past information, aiding P-NTM's control mechanism, which would otherwise only consider the current token.
All models share a common structure consisting of an initial token-embedding layer and a final softmax layer for computing output token probabilities.
They are also designed to have comparable parameter counts.

\paragraph{Training.}
Models are trained on randomly sampled batches of problems for up to 500{,}000 iterations.
At each iteration, an input length $\ell \in [1, 40]$ is chosen uniformly at random, and 128 examples of that length are drawn.
Each training example consists of the input and output sequences concatenated with the separator token.
Training uses supervised learning with cross-entropy loss on the output tokens, conditioning each prediction on the previous ground-truth token.
The Adam optimizer~\citep{kingma2017adam} is used, with the learning rate fixed at $5 \times 10^{-4}$ for the entire duration of training.
Early stopping is triggered if the infinity norm of the parameter gradient remains below $10^{-8}$ for 500 consecutive iterations.
To assess sensitivity to initialization, ten models of each architecture are trained per task, each with a different random seed.

\paragraph{Evaluation.}
Each trained model is evaluated using greedy decoding to generate outputs from the inputs.
While more sophisticated decoding strategies could yield better results, we adopt greedy decoding as a simple baseline.
Generalization performance is measured using \emph{exact match accuracy} on problems with input lengths $\ell \in [41, 120]$, corresponding to inputs up to three times longer than those observed during training.
For each value of $\ell$, accuracy is estimated using 128 randomly sampled inputs and is computed as the proportion of generated output sequences that exactly match the target sequence.
For models with external memory (NTM and P-NTM), the memory size is increased relative to the training configuration to accommodate the larger problem sizes used during evaluation.

\subsection{Computational Efficiency}
To evaluate computational efficiency, the experiment measures how inference time scales with input length for both a standard NTM and the proposed P-NTM.
For P-NTM, we measure both sequential and parallel execution times to isolate the impact of parallelization from other architectural simplifications.
The goal is to quantify the speed advantage offered by parallel computation, particularly for long sequences.

The NTM model in this experiment uses an LSTM controller, while the P-NTM model employs a preceding minGRU layer, following the architectures described in the previous section.
Both models use a single read–write head pair, a hidden dimension of 128, and a fixed memory size of 512.
To match the parameter count of the LSTM controller, the minGRU layer in the P-NTM model uses a state expansion factor of 3.

Input lengths ranging from $2^3 = 8$ to $2^{16} = 65{,}536$ are tested, increasing in powers of two.
For each length, a batch of 8 synthetic sequences of 128-dimensional vectors is generated, and inference latency is measured over the batch.
Then, 10 timed runs are performed to obtain the runtime measurements, preceded by 3 warm-up runs to account for compilation and caching effects.
Finally, the mean speedup of P-NTM relative to NTM is reported for both sequential and parallel execution, calculated as the ratio of their respective mean runtimes.
All measurements are obtained on an NVIDIA A100 system, providing a high-throughput environment for runtime comparisons.

\section{Results}
\label{sec:results}

\subsection{Length Generalization}

\begin{table}
	\centering
	\caption{Maximum exact match accuracy scores across runs for each architecture and task on problems of unseen length ($\ell = 41$ to $\ell = 120$). Perfect scores are highlighted in bold.}
	\begin{tabular}{lcccccc}
		\hline
		  \textbf{Architecture}          & \textbf{PC}              & \textbf{CN}              & \textbf{RS}              & \textbf{DS}              & \textbf{MA}              & \textbf{BA}              \\
		\hline
		LSTM         & 0.09            & 0.04            & 0.14            & 0.09            & 0.06            & 0.12            \\
		minGRU       & 0.07            & 0.04            & 0.09            & 0.09            & 0.08            & 0.10            \\
		Transformer  & 0.27            & 0.56            & 0.21            & 0.65            & 0.42            & 0.22            \\
		NTM          & $\mathbf{1.00}$ & $\mathbf{1.00}$ & $\mathbf{1.00}$ & $\mathbf{1.00}$ & $\mathbf{1.00}$ & $\mathbf{1.00}$ \\
		P-NTM     & $\mathbf{1.00}$ & $\mathbf{1.00}$ & $\mathbf{1.00}$ & $\mathbf{1.00}$ & $\mathbf{1.00}$ & $\mathbf{1.00}$ \\
		\hline
	\end{tabular}
	\label{tab:max-scores}
\end{table}

\Cref{tab:max-scores} reports the maximum exact match accuracy achieved across runs for each architecture\footnote{For Transformers, only the highest-scoring positional encoding variant is shown. Full results are provided in Appendix~\ref{ap:additional-results}.} on each task when evaluated on sequences of unseen length.
This metric indicates whether any training run found a parameter configuration capable of extrapolating beyond the training data and solving instances substantially larger than those observed during training.

The results show that, like the standard NTM, P-NTM successfully learns all evaluated computational tasks, with both architectures achieving perfect accuracy across all settings.
In contrast, the recurrent baselines (LSTM and minGRU) perform poorly, with accuracy remaining below 0.12 in all cases.
Transformer models generalize better than recurrent networks but still fall short of the Turing-inspired architectures, achieving at most 0.65 accuracy.

\subsection{Sensitivity to Initialization}

\begin{figure}[t]
	\centering
	\includegraphics[width=0.9\linewidth]{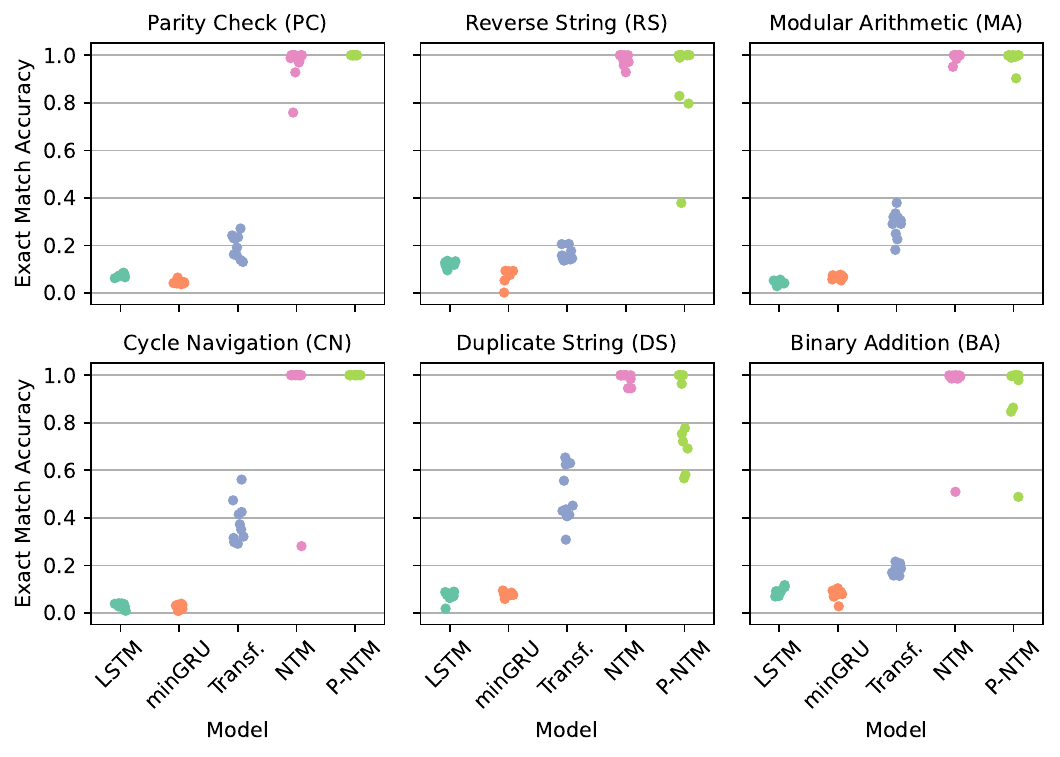}
	\caption{Exact match accuracy scores across runs for different architectures and tasks on problems of unseen length ($\ell = 41$ to $\ell = 120$).}
	\label{fig:sensitivity}
\end{figure}

\Cref{fig:sensitivity} shows, for each task, the distribution of exact match accuracies obtained under different initialization seeds.
This highlights variability in learning outcomes and reveals how reliably each architecture converges to solutions that generalize to unseen sequence lengths.

Both NTM and P-NTM occasionally achieve perfect generalization across all evaluated lengths, but not consistently.
Specifically, inspection of the per-run accuracies in the figure shows that perfect accuracy is attained in 35\% of runs for NTM and 55\% for P-NTM, suggesting an advantage for the latter in discovering fully generalizing parameter configurations.
However, when performance degrades, P-NTM tends to fail more sharply, exhibiting a lower mean accuracy (0.93 vs. 0.97) and a larger standard deviation (0.14 vs. 0.11), indicating reduced stability across runs.

By contrast, recurrent architectures largely fail to generalize to longer lengths, while Transformers exhibit mixed behavior: some runs achieve partial generalization, but none match the consistency of the memory-augmented models.

\subsection{Speedup}

\begin{figure}[t]
	\centering
	\includegraphics[scale=0.85]{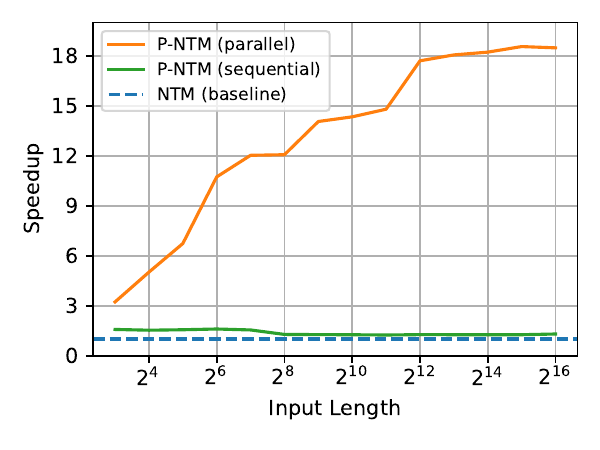}
	\caption{Speedup of P-NTM sequential and parallel execution relative to the standard NTM execution across input batches of varying sequence lengths.}
	\label{fig:efficiency}
\end{figure}

\Cref{fig:efficiency} shows the speedup achieved by both parallel and sequential P-NTM execution relative to the standard NTM as input length increases.
Sequential P-NTM execution attains a speed comparable to that of the standard NTM across all input lengths.
In contrast, parallel P-NTM consistently outperforms the baseline, with inference running $3.6\times$ to $18.5\times$ faster than the standard NTM, and speedup increasing with input length.
The timings for each architecture and input length are provided in Appendix~\ref{ap:additional-results}.

%% file: sections/05_discussion.tex
\section{Discussion}
\label{sec:discussion}

The experimental results support the hypothesis that P-NTM preserves the computational expressiveness of the standard NTM despite removing an explicit recurrent controller state.
P-NTM successfully learns all evaluated algorithmic tasks and generalizes to significantly longer sequences, matching the performance of the original NTM.
This indicates that explicit controller state is not required to represent finite-state computation when external memory is available and control can be mediated autoregressively through output tokens, and that this strategy can be learned from data.

Transformer models outperform simple recurrent baselines, reflecting their higher capacity and more flexible access to past information.
However, they remain less effective than the Turing-inspired architectures on these tasks and fail to reliably generalize to longer, unseen sequences.
This behavior is consistent with prior findings on Transformer length generalization on algorithmic tasks~\citep{deletang2023chomsky,zheng2024dape}.

Differences between NTM and P-NTM appear in their sensitivity to initialization.
Although P-NTM reaches perfect generalization more often, it also shows greater variability across runs.
This is because failures lead to more abrupt accuracy degradation, a pattern that is especially apparent when accuracy is plotted against sequence length (see Appendix~\ref{ap:additional-results}).
This variability may stem from the simplified stability mechanism adopted by P-NTM, where the threshold-based control of shift weights may not fully prevent errors from accumulating.

Finally, the measured speedups highlight the substantial benefits of parallelism.
While the simplifications in P-NTM give it a modest performance advantage over the standard NTM, the results indicate that parallelism accounts for the majority of the speedup.
This advantage, however, comes with a tradeoff: standard NTM inference uses constant memory per step, whereas parallel P-NTM execution via scan requires memory that grows linearly with sequence length.
As a result, full parallelization is limited by available memory, and very long sequences may require chunked execution.

%% file: sections/06_limitations.tex
\section{Limitations}
\label{sec:limitations}

A central tradeoff of P-NTM is its reliance on sequential interaction with output tokens to represent and update computational state.
This design performs well during training when problems are paired with sufficiently detailed intermediate reasoning steps, enabling the model to learn these computations in parallel.
But when such intermediate outputs are missing, sparse, or noisy, the model may struggle to reliably learn and track discrete states~\citep{merrill2024illusion}.
In contrast, standard NTMs maintain state internally within a recurrent controller and can, in principle, learn effective state-tracking behavior even in the absence of intermediate supervision, such as when training data consists of blank or uninformative tokens.

One potential direction for addressing this limitation is to explore augmenting P-NTM with more expressive linear recurrent architectures.
Recent work on the DeltaNet family of models suggests that such architectures can represent and learn regular languages while remaining parallelizable~\citep{grazzi2025unlocking,yang2024delta,siems2025deltaproduct}.
Incorporating similar components into P-NTM could allow the model to retain the parallelism benefits of its design while introducing an implicit finite control state, akin to that of standard NTMs.
This, in turn, may reduce reliance on explicit intermediate reasoning steps in the training data.

At the same time, the benefits of P-NTM are limited in settings where computation is inherently sequential.
In autoregressive generation regimes, including some online reinforcement learning scenarios, per-step computation cost is typically the dominant bottleneck.
Because generation is inherently sequential, the parallelism advantages of P-NTM during training do not carry over to inference-time execution.
In these regimes, scalability is instead limited by the need to maintain a larger memory matrix as problem size increases, along with the cumulative cost of executing autoregressive generation steps.

More broadly, these limitations highlight the challenge of designing architectures that balance computational expressiveness with efficient sequential inference.
In addition to architectural augmentation, future work could explore more compact memory representations, improved addressing mechanisms, or controller designs that lower the cost of each recurrent update while preserving expressive power.
Advances along these lines could improve scalability across a wider range of learning and inference regimes.

%% file: sections/07_related_work.tex
\section{Related Work}
\label{sec:relwork}

\paragraph{Turing-inspired neural networks.}
Our work is related to several extensions of the original NTM architecture.
Differentiable Neural Computer (DNC)~\citep{graves2016dnc} introduces additional memory management mechanisms, including a temporal link matrix and memory usage tracking, which together enable more flexible location-based access patterns and mitigate write interference.
Dynamic Neural Turing Machine (D-NTM)~\citep{gulcehre2017dntm}, on the other hand, aims to improve the flexibility of memory access patterns by introducing fixed, learnable address vectors and a recency-based access strategy, with both continuous and discrete variants.
Meanwhile, our work is orthogonal to these extensions and instead focuses on parallelization, rather than on increasing memory flexibility or stability.
More similar to our work is Baby-NTM~\citep{suzgun2019baby}, which simplifies the NTM by removing content-based addressing and retaining only location-based addressing.
Unlike Baby-NTM, however, our method introduces a targeted simplification of the NTM that enables parallel computation.
More broadly, Neural GPUs~\citep{kaiser2016gpu} pursue a goal similar to ours and seek to address the sequential bottleneck of NTMs.
They do so by removing explicit memory operations and instead iteratively applying convolutional layers over the entire input sequence to produce the output.
While this design enables fully parallel computation, it does not naturally support recurrent execution for autoregressive generation.
In contrast, our approach preserves recurrent execution while still enabling parallel computation.
Finally, nnTM~\citep{stogin2024nntm} is a Turing-complete architecture based on higher-order connections and stack-based memory, with provable inference-time stability guarantees.
Instead, we retain an explicit memory matrix with moving heads, as in NTMs, while adopting a parallelizable architecture.
Inference stability is encouraged through discrete autoregressive outputs, as well as clipping and renormalization of the addressing weights.

\paragraph{Parallelizable recurrent networks.}
Recent work has focused on designing recurrent neural networks that support parallel training while retaining efficient sequential inference, often by modeling sequential dependencies through linear recurrence relations.
Early studies explored recurrent networks with linear recurrent updates~\citep{bradbury2017quasirecurrent,lei2018sru} and showed that such models can be parallelized using scan algorithms~\citep{martin2018rnn}.
Subsequent work drew inspiration from state-space models to develop linear recurrent architectures~\citep{gu2020hippo,gu2022s4,gu2022s4d,smith2023s5}, leading to the Mamba family of models~\citep{gu2024mamba,dao2024mamba2}, which achieve strong efficiency through scan parallelism, linear-time inference, and competitive performance in language modeling.
Architectures not directly tied to state-space formulations have also been proposed, such as the Linear Recurrent Unit (LRU)~\citep{orvieto2023lru}, while more recent work extends parallelization techniques to more complex recurrences involving memory updates based on gradient optimization steps, including Titans~\citep{behrouz2024titans} and DeltaNet~\citep{yang2024delta,yang2025gated}.
Most closely related to our work, \citet{feng2024rnns} revisited LSTM~\citep{hochreiter1997lstm} and GRU~\citep{cho2014gru}, introducing simplified variants named minLSTM and minGRU that remove temporal dependencies in state updates and enable scan parallelization.
Our work extends these ideas to the NTM architecture.

\paragraph{Expressiveness of neural sequence models.}
Several works have studied the expressive power of sequence modeling architectures, particularly those used by language models, by relating them to formal language classes and classical models of computation.
It is well established that standard recurrent neural networks correspond to finite automata and thus recognize regular languages~\citep{siegelmann1996finite,horne1996rnn,weiss2018practical}.
More recently, research has examined linear recurrent architectures~\citep{merrill2024illusion,sarrof2024the}, showing that specific classes of such models can efficiently represent and exactly recognize regular languages~\citep{grazzi2025unlocking,siems2025deltaproduct}.
Transformers, in contrast, have been shown to have limited state-tracking capabilities, as their expressive power is restricted to constant-depth circuit classes~\citep{merrill2023parallelism}.
However, subsequent work has shown that autoregressive chain-of-thought generation can overcome many of these limitations~\citep{feng2023mystery,li2024serial}.
In particular, chain-of-thought transformers have been shown to learn and represent state tracking~\citep{amiri2025lower,huang2025cot} and even to achieve universal computation through autoregressive generation~\citep{jiang2025turing,schuurmans2024universal}, though these results do not always generalize across sequence lengths, may rely on specialized training objectives, or assume structured autoregressive generation formats that are not typically reflected in training data.
Our work draws inspiration from these perspectives and adopts a practical focus.
Rather than characterizing theoretical limits, we contribute an autoregressive architecture that aims to balance expressiveness with efficiency in training and inference, while supporting generalization to unseen sequence lengths.
We present a concrete, parallelizable architecture inspired by a classical model of computation and provide an empirical evaluation of its behavior.

%% file: sections/08_conclusion.tex
\section{Conclusion}
\label{sec:conclusion}

We introduce Parallelizable Neural Turing Machine (P-NTM), a simplified variant of the original NTM that enables efficient parallel computation via scan algorithms while preserving the expressiveness required for algorithmic tasks.
On synthetic algorithmic benchmarks, P-NTM successfully learns state tracking, memorization, and arithmetic, and generalizes to sequences longer than those seen during training, matching the performance of standard NTMs while achieving speedups of an order of magnitude on long inputs.
Thus, P-NTM represents a contribution toward better balancing computational expressiveness with training efficiency.

\section*{Acknowledgements}

This study was financed in part by the Coordenação de Aperfeiçoamento de Pessoal de Nível Superior---Brasil (CAPES)---Finance Code 001.

%% file: appendix/a.tex
\section{Experimental Details}
\label{ap:experimental-details}

\subsection{Algorithm Learning and Representation}

\subsubsection{Task Definitions}
\label{aps:task-definitions}

\paragraph{Parity Check.}
Consists in taking a binary string composed of $a$s and $b$s and determining whether the number of $a$s is odd or even.
Given an input $x \in \{a,b\}^n$, the goal is to produce the output $y \in \{0,1\}^n$ where the $k$th symbol $y_k$ is the parity of the count of $a$s in the prefix $x_{1}\dots x_k$.  
The final symbol $y_n$ therefore encodes the parity of the entire string.

\paragraph{Cycle Navigation.}
Consists in simulating movement on a 5-state cycle using operations for staying put, moving forward (i.e., increment), or moving backward (i.e., decrement).
Let $\mathcal S = \{0,\dots,4\}$.
The input is $x \in \{s,i,d\}^n$, where $s$ denotes ``stay,'' (i) denotes ``increment,'' and $d$ denotes ``decrement,'' all with wrap-around at the boundaries.
The output is $y \in \mathcal S^n$, where each $y_k$ corresponds to the state reached after applying operations $x_1,\dots,x_k$ starting from state 0.

\paragraph{Reverse String.}
Consists in reversing the input string.
For $x = x_1 \dots x_n \in \{0,1\}^n$, the output is $y = x_n \dots x_1$.

\paragraph{Duplicate String.}
Consists in repeating the input string twice with no separator.
Given $x = x_1 \dots x_n \in \{0,1\}^n$, the output is $y = xx = x_1 \dots x_n x_1 \dots x_n$.

\paragraph{Modular Arithmetic.}
Consists in evaluating a modular arithmetic expression involving addition, subtraction, and multiplication modulo $5$.
Given an input expression that interleaves operators in $\{+,-,\times\}$ with integers in \{0,1,2,3,4\} (e.g., $1 \times 2 + 4$),\footnote{Because expressions alternate operand and operator tokens, inputs must have odd length. If an even length is requested, we generate an input one token longer instead.} the goal is to compute its value by maintaining partial products and partial results while scanning the expression from left to right in a single pass.
To this end, the output assumes the form $y = s_1 \dots s_n v$, where each $s_i$ is a triple of symbols $h_1 h_2 h_3$ representing an intermediate step, and $v$ is the final result of the expression modulo $5$.
Here, $h_1$ encodes the sign of the current partial product, $h_2$ encodes the value of this partial product modulo $5$, and $h_3$ encodes the current partial result of the expression (also modulo $5$).
For example, the triple $+12$ indicates that the partial product is positive with value $1$, and the running value of the expression at that step is $2$.

\paragraph{Binary Addition.}
Consists in adding two binary numbers written in little-endian order and outputting their sum, also in little-endian form.
The input is a string of the form $x = b_1 + b_2$, where $b_1 \in \{0,1\}^m$ and $b_2 \in \{0,1\}^n$.
The output is $y = \operatorname{bin}(\operatorname{int}(b_1) + \operatorname{int}(b_2))$, namely the binary string representation of the integer sum, again written in little-endian order.

\subsubsection{Architecture Descriptions}

\begin{figure}[t]
    \centering
    \includegraphics[width=0.35\linewidth]{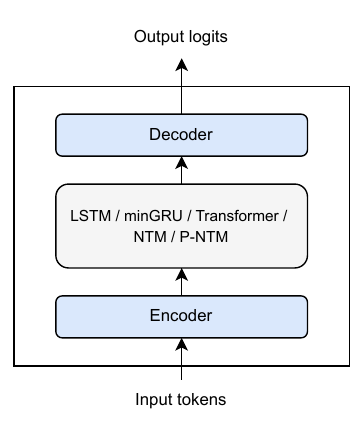}
    \caption{High-level structure shared by all evaluated models. An encoder maps input tokens to embeddings, a sequence-processing module transforms them, and a decoder produces vocabulary logits.}
    \label{fig:general-structure}
\end{figure}

All models used in the experiment share the same high-level structure, illustrated in \Cref{fig:general-structure}.
Each model consists of (i) an encoder that maps input tokens to embeddings, (ii) a sequence-processing module, and (iii) a decoder that produces logits over the vocabulary via linear projection.
The only component that differs between models is the inner sequence-processing module, which corresponds to the underlying architecture under investigation.
Detailed configurations for each architecture are provided below, and \Cref{tab:model_params} summarizes the corresponding hyperparameters.

\paragraph{LSTM.}
Consists of a single LSTM layer applied to the embedded token sequence, with both the embedding dimension and the LSTM hidden size set to $192$.
The LSTM produces a sequence of hidden states, which are linearly projected by a decoder layer to generate output logits at each timestep.

\begin{figure}[t]
    \centering
    \includegraphics[width=0.4\textwidth]{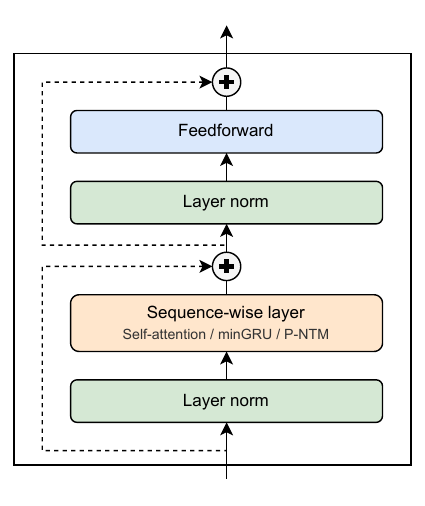}
    \caption{Residual block structure used by the minGRU, Transformer, and P-NTM models. Each block applies a sequence-wise sublayer (self-attention, minGRU, or P-NTM layer) over the full sequence, followed by a position-wise feedforward network. Both sublayers use pre-normalization and residual connections. LSTM and NTM architectures do not adopt this structure.}
    \label{fig:sequencewise-block}
\end{figure}

\paragraph{minGRU.}
Comprises a stack of $5$ identical residual minGRU blocks with hidden size $64$.
Each block includes (i) a pre-normalized minGRU sequence-mixing sublayer with a state-expansion factor of $2$ (yielding a state size of $128$), followed by (ii) a pre-normalized position-wise feedforward sublayer with $4\times$ expansion and GELU activation.
Both sublayers are wrapped in residual connections.
See \Cref{fig:sequencewise-block}.

\paragraph{Transformer.}
Consists of $5$ identical residual Transformer decoder blocks with embedding width $64$.
Each block contains (i) a pre-normalized multi-head causal self-attention sublayer with $8$ heads, followed by (ii) a pre-normalized position-wise feedforward sublayer with $4\times$ expansion and GELU activation; both sublayers use residual connections.
Positional information is incorporated using one of the following methods: no explicit positional encoding (NoPE), RoPE (rotary embeddings with $\theta = 10{,}000$), ALiBi, or FIRE (using a learned bias network with hidden width $32$, initialized with $c_0 = 0.1$ and $L_0 = 512$, and $\varepsilon = 10^{-6}$).
See \Cref{fig:sequencewise-block}.

\paragraph{NTM.}
Consists of an NTM controller with embedding width $104$ and an external memory composed of cells of size $32$.
At each timestep, the controller interacts with memory through $4$ read heads and $4$ write heads, producing a sequence of output representations.
These representations are projected to token logits by a decoder layer.

\paragraph{P-NTM.}
Consists of two residual blocks with embedding width $104$.
The first block is a minGRU block as described above.
The second block is a P-NTM block, whose sequence-mixing sublayer comprises a P-NTM layer with memory cell size $32$ and $4$ heads, followed by a position-wise feedforward sublayer with $4\times$ expansion and GELU activation.
All sublayers are pre-normalized and enclosed within residual connections.
See \Cref{fig:sequencewise-block}.

\begin{table}[t]
    \centering
    \caption{Hyperparameters and parameter counts for all evaluated models. Dashes (---) indicate the hyperparameter is not applicable to that specific architecture.}
    \label{tab:model_params}
    \begin{tabular}{lccccc}
        \toprule
        \textbf{Architecture} & \textbf{Hidden Size} & \textbf{Depth} & \textbf{Heads} & \textbf{Cell Size} & \textbf{Parameters} \\
        \midrule
        LSTM & 192 & --- & --- & --- & $\sim$298 K \\
        minGRU & 64 & 5 & --- & --- & $\sim$289 K \\
        Transformer (NoPE) & 64 & 5 & 8 & --- & $\sim$248 K \\
        Transformer (ALiBi) & 64 & 5 & 8 & --- & $\sim$248 K \\
        Transformer (RoPE) & 64 & 5 & 8 & --- & $\sim$248 K \\
        Transformer (FIRE) & 64 & 5 & 8 & --- & $\sim$249 K \\
        NTM & 104 & --- & 4 & 32 & $\sim$224 K \\
        P-NTM & 104 & --- & 4 & 32 & $\sim$260 K \\
        \bottomrule
    \end{tabular}
\end{table}

\subsubsection{Training Configuration}

During training, both NTM and P-NTM used a memory size of $m_\mathrm{train} = 2 \times \ell_{\max} + 16 = 96$, where $\ell_{\max}$ is the maximum input length observed during training. 
This setting was chosen to avoid positional interference resulting from circular wraparound of the write heads.

\subsubsection{Evaluation Configuration}

During evaluation, the memory size for both NTM and P-NTM was increased to $256$. 
As in training, memory was allocated to be twice the maximum input length, with additional slack. 
For P-NTM, a stability threshold of $\tau = 0.01$ was used.

\subsubsection{Random Seeds}

The random seeds were generated using NumPy's random number generator, starting from an initial seed of 0 and producing 10 derived seed values.

\subsection{Computational Efficiency}

\paragraph{Model size.} 
For the computational efficiency experiments, the standard NTM model contains 168{,}140 trainable parameters. The P-NTM model contains 152{,}576 parameters in total, of which 147{,}456 correspond to the minGRU controller.

\paragraph{Hardware configuration.} 
The experiments were conducted on a single NVIDIA A100 SXM4 GPU (80\,GB HBM2e, 1522.0\,GB/s memory bandwidth, CUDA 12.x). The host system was equipped with an AMD EPYC 7532 CPU (64 logical cores) and 129\,GB of RAM.

%% file: appendix/b.tex
\section{Additional Results}
\label{ap:additional-results}

\subsection{Length Generalization}

\begin{table}
	\centering
	\caption{Maximum exact match accuracy scores across runs for each task and Transformer positional encoding scheme. Best results boldfaced. }
	{
        \begin{tabular}{lcccccc}
        \hline
        \textbf{Pos. Encoding} & \textbf{PC} & \textbf{CN} & \textbf{RS} & \textbf{DS} & \textbf{MA} & \textbf{BA} \\
            \hline
            NoPE & 0.16 & 0.19 & 0.18 & 0.24 & 0.19 & \textbf{0.22} \\
            RoPE & 0.06 & 0.07 & 0.05 & 0.05 & 0.04 & 0.06 \\
            ALiBi & \textbf{0.27} & \textbf{0.56} & \textbf{0.21} & \textbf{0.65} & 0.38 & 0.20 \\
            FIRE & 0.23 & 0.44 & 0.19 & 0.28 & \textbf{0.42} & 0.19 \\
            \hline
        \end{tabular}
    }
	\label{tab:transformer-results}
\end{table}

\Cref{tab:transformer-results} compares the length generalization performance of the Transformer architecture under different positional encoding schemes on the algorithmic learning and representation tasks.
ALiBi is the most robust, achieving the highest accuracy in most settings, with FIRE close behind and occasionally ranking first.
RoPE generalizes poorly beyond the training length, performing near the lower bound, while NoPE is competitive in one task but overall trails ALiBi and FIRE.

\begin{figure}[ht]
    \centering
    \includegraphics[scale=0.75]{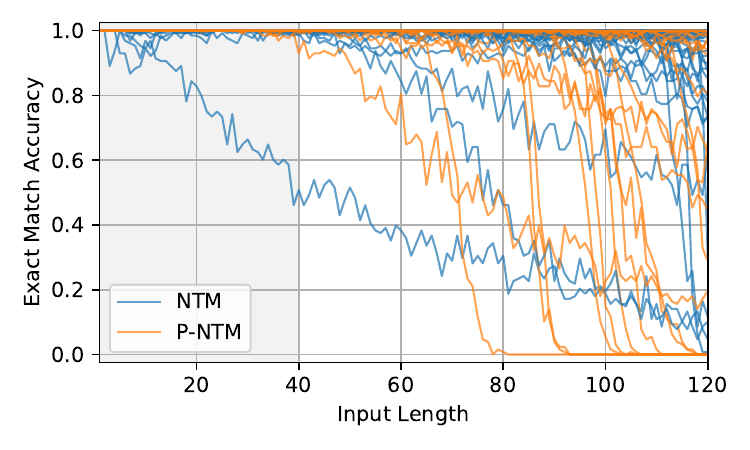}
    \caption{Exact match accuracy across varying input lengths for NTM and P-NTM. Each curve represents a separate run on a specific task, with the shaded area indicating the training distribution ($\ell = 1$ to $\ell = 40$). While NTM generally maintains non-zero accuracy at extended lengths, P-NTM exhibits a sharp performance degradation, frequently dropping to zero accuracy for longer inputs.}
    \label{fig:acclength}
\end{figure}

\Cref{fig:acclength} shows the generalization performance of the NTM and P-NTM models as sequence lengths extend beyond the training distribution.
Both architectures maintain near-perfect exact match accuracy within and just beyond the training bounds (shaded region).
However, for input lengths greater than 40, performance becomes unstable and begins to deteriorate.
The key difference appears in their failure modes: P-NTM exhibits sharp, catastrophic drops in performance, with many runs collapsing to zero accuracy between lengths 70 and 120.
In contrast, NTM degrades more gradually and typically retains partial, non-zero accuracy across the full evaluated range.

\subsection{Computational Efficiency}

\begin{table}[ht]
\centering
\caption{Mean execution time ($\pm$ standard deviation), measured in seconds, across batches of input sequences of varying lengths for NTM and P-NTM. The P-NTM results include both sequential and parallel execution.}
\begin{tabular}{lccc}
\hline
\textbf{Input Length} & \textbf{NTM} & \textbf{P-NTM (Sequential)} & \textbf{P-NTM (Parallel)} \\ \hline
$2^{3} = 8$ & $0.0011 \pm 0.0000$ & $0.0007 \pm 0.0000$ & $0.0003 \pm 0.0000$ \\ 
$2^{4} = 16$ & $0.0019 \pm 0.0001$ & $0.0012 \pm 0.0000$ & $0.0004 \pm 0.0000$ \\
$2^{5} = 32$ & $0.0035 \pm 0.0000$ & $0.0023 \pm 0.0000$ & $0.0005 \pm 0.0000$ \\ 
$2^{6} = 64$ & $0.0068 \pm 0.0001$ & $0.0043 \pm 0.0000$ & $0.0006 \pm 0.0000$ \\ 
$2^{7} = 128$ & $0.0128 \pm 0.0000$ & $0.0083 \pm 0.0000$ & $0.0011 \pm 0.0001$ \\
$2^{8} = 256$ & $0.0209 \pm 0.0003$ & $0.0164 \pm 0.0000$ & $0.0017 \pm 0.0002$ \\ 
$2^{9} = 512$ & $0.0411 \pm 0.0001$ & $0.0325 \pm 0.0001$ & $0.0029 \pm 0.0000$ \\
$2^{10} = 1,024$ & $0.0818 \pm 0.0001$ & $0.0651 \pm 0.0008$ & $0.0057 \pm 0.0001$ \\
$2^{11} = 2,048$ & $0.1620 \pm 0.0001$ & $0.1305 \pm 0.0020$ & $0.0109 \pm 0.0001$ \\
$2^{12} = 4,096$ & $0.3268 \pm 0.0001$ & $0.2588 \pm 0.0007$ & $0.0185 \pm 0.0000$ \\
$2^{13} = 8,192$ & $0.6546 \pm 0.0004$ & $0.5197 \pm 0.0014$ & $0.0363 \pm 0.0000$ \\
$2^{14} = 16,384$ & $1.3079 \pm 0.0001$ & $1.0396 \pm 0.0017$ & $0.0718 \pm 0.0001$ \\
$2^{15} = 32,768$ & $2.6242 \pm 0.0019$ & $2.0857 \pm 0.0030$ & $0.1414 \pm 0.0001$ \\
$2^{16} = 65,536$ & $5.2328 \pm 0.0018$ & $4.0207 \pm 0.0200$ & $0.2832 \pm 0.0001$ \\
\hline
\end{tabular}
\label{tab:timings}
\end{table}

\Cref{tab:timings} reports the mean execution times in seconds for the NTM and P-NTM variants across exponentially increasing input lengths from $2^3$ to $2^{16}$.
The standard NTM and the sequential P-NTM exhibit similar scaling behavior, with the sequential execution of P-NTM providing a consistent but modest reduction in latency, amounting to approximately 1.2 seconds at the largest input size.
In contrast, the parallel execution of P-NTM shows a markedly different scaling pattern.
It processes a sequence of length $2^{16}$, or 65{,}536 elements, in 0.2832 seconds, whereas the standard NTM requires 0.3268 seconds to process a sequence of length $2^{12}$, or 4{,}096 elements.
Thus, within a comparable time budget, the parallel architecture accommodates a sixteen-fold increase in sequence length while maintaining sub-second execution time.

\subsection{Learned Patterns}

\begin{figure}[ht]
    \centering
    \includegraphics[width=\linewidth]{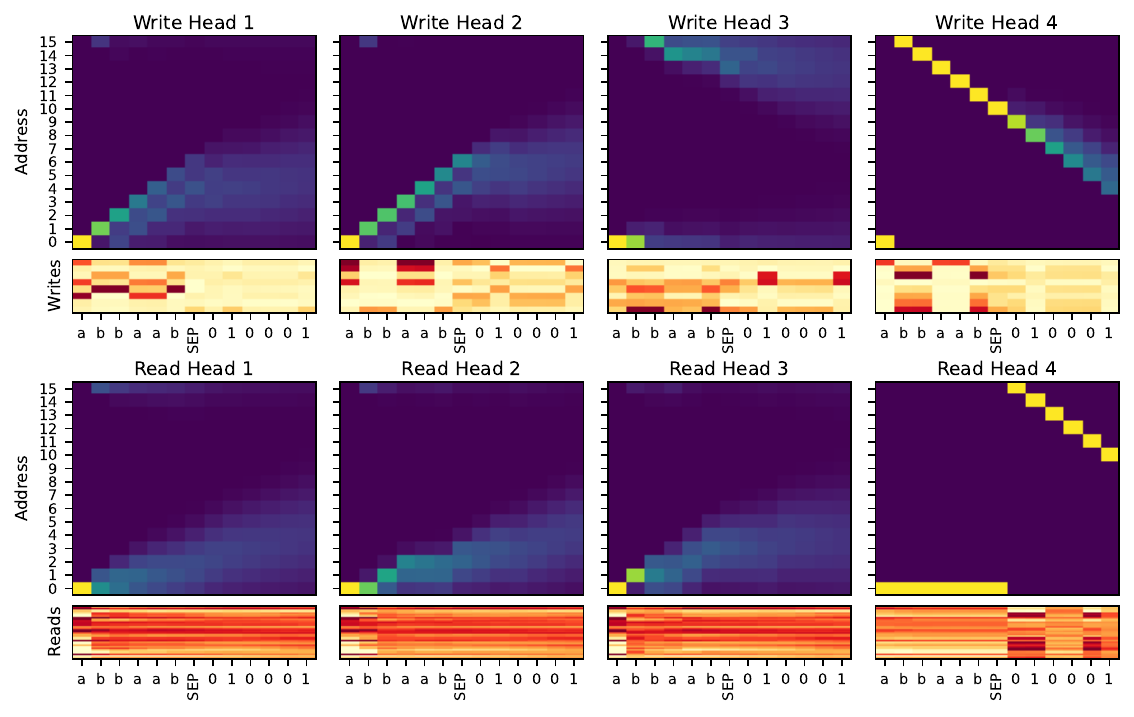}
    \caption{Memory interaction patterns of the heads of a trained P-NTM on a parity check sequence. The top row displays \textbf{write heads}, and the bottom row displays \textbf{read heads}. For each head, the upper panel visualizes addressing weights across memory locations over time, while the lower strip shows the read or write vector at each timestep (normalized per head). Write Head 4 and Read Head 4 exhibit the sharpest addressing weights.}
    \label{fig:memory}
\end{figure}

\Cref{fig:memory} illustrates the memory access patterns and read–write dynamics of a trained P-NTM model as it processes and solves a parity check instance.
Among the heads, Write Head 4 and Read Head 4 seem to play a particularly central role.
While processing the input, write Head 4 advances monotonically through memory, incrementing the address for each input token to store distinct symbol representations in consecutive locations, with write vectors shifting in sync with the input sequence.
During output generation, Read Head 4 retrieves information from these same memory locations in their original order, with read vectors reflecting the stored input rather than the current output token.
This behavior suggests that the model solves the parity task by first buffering the input sequence into memory and then sequentially re-reading it to generate the output.

\clearpage

%% file: appendix/c.tex
\section{Complete Parallel Execution Algorithm}

\begin{algorithm}[H]
	\caption{$\bm y_1,\dots,\bm y_T \gets \texttt{MultiHeadPNTM}(\bm x_1, \dots, \bm x_T \mid \bm{\mathcal{W}})$}
	\begin{algorithmic}[1]
		\Statex \textbf{Input:} $\bm x_1, \dots, \bm x_T$, where $\bm x_t \in \mathbb{R}^{d}$
		\Statex \textbf{Output:} $\bm y_1, \dots, \bm y_T$, where $\bm y_t \in \mathbb{R}^{d}$
		\Statex \textbf{Hyperparameters:} $H$, number of heads
		\Statex \textbf{Parameters:} $\bm{\mathcal{W}}$, consisting of
		\Statex \hspace{\algorithmicindent}
		$\bm W_r^{h} \in \mathbb{R}^{3 \times d}$,
		$\bm W_w^{h} \in \mathbb{R}^{3 \times d}$,
		$\bm W_u^{h} \in \mathbb{R}^{(n/H) \times d}$ for $h = 1,\dots,H$;
		\Statex \hspace{\algorithmicindent}
		$\bm W_m \in \mathbb{R}^{n \times n}$ and
		$\bm W_o \in \mathbb{R}^{d \times (H n)}$

		\ParFor{$h \gets 1 \textbf{ to } H$}
		\ParFor{$t \gets 1 \textbf{ to } T$} \Comment{Control}
			\State $\bm s_{t}^{r,h} \gets \softmax(\bm W_r^{h} \bm x_t)$
			\State $\bm s_{t}^{w,h} \gets \softmax(\bm W_w^{h} \bm x_t)$
			\State $\bm u_{t}^{h} \gets \bm W_u^{h} \bm x_t$
		\EndParFor

		\State $\bm a_{0}^{r,h}, \dots, \bm a_{T-1}^{r,h}
		\gets \texttt{ConvShift}(\bm s_{1}^{r,h}, \dots, \bm s_{T}^{r,h})$
		\Comment{Addressing}

		\State $\bm a_{0}^{w,h}, \dots, \bm a_{T-1}^{w, h}
		\gets \texttt{ConvShift}(\bm s_{1}^{w,h}, \dots, \bm s_{T}^{w,h})$

		\State $\bm M_{1}^{h}, \dots, \bm M_{T}^{h}
		\gets \texttt{MemoryWrite}(\bm a_{0}^{w,h}, \dots, \bm a_{T-1}^{w,h},
		\bm u_{1}^{h}, \dots, \bm u_{T}^{h})$
		\Comment{Writing}
		\EndParFor

        \ParFor{$t \gets 1 \textbf{ to } T$}
            \State $\tilde{\bm M}_t \gets \bm W_m \big[ \bm M_t^{1\intercal} \ldots \bm M_t^{H\intercal} \big]^{\intercal}$
            \Comment{Mixing}
        
            \State $\bm r_t \gets
            \begin{bmatrix}
                \tilde{\bm M}^{\intercal}_t \bm a_{t-1}^{r,1} &
                \cdots &
                \tilde{\bm M}^{\intercal}_t \bm a_{t-1}^{r,H}
            \end{bmatrix}$
            \Comment{Reading}
        
            \State $\bm y_t \gets \bm W_o \bm r_t$
            \Comment{Output}
        \EndParFor

		\State \Return $\bm y_1,\dots,\bm y_T$
	\end{algorithmic}
\end{algorithm}